\documentclass[runningheads]{llncs}
\usepackage{amsmath,amsfonts,bm}

\def\eqref#1{equation~\ref{#1}}
\def\1{\bm{1}}

\newcommand{\test}{\mathcal{D_{\mathrm{test}}}}

\def\mF{{\bm{F}}}

\DeclareMathAlphabet{\mathsfit}{\encodingdefault}{\sfdefault}{m}{sl}
\SetMathAlphabet{\mathsfit}{bold}{\encodingdefault}{\sfdefault}{bx}{n}
\newcommand{\tens}[1]{\bm{\mathsfit{#1}}}
\def\tA{{\tens{A}}}
\def\tB{{\tens{B}}}
\def\tC{{\tens{C}}}

\def\tG{{\tens{G}}}

\def\tP{{\tens{P}}}
\def\tQ{{\tens{Q}}}
\def\tR{{\tens{R}}}

\def\tW{{\tens{W}}}
\def\tX{{\tens{X}}}

\def\sR{{\mathbb{R}}}

\usepackage{graphicx}
\usepackage{comment}
\usepackage{amsmath,amssymb} % define this before the line numbering.
\usepackage{color}
\usepackage{multirow}
\usepackage{subfig}
\usepackage{booktabs}

\definecolor{8}{rgb}{0.973,0.412,0.42}
\definecolor{7}{rgb}{0.961,0.411,0.419}
\definecolor{6}{rgb}{0.980,0.54,0.45}
\definecolor{5}{rgb}{0.988,0.666,0.4705}
\definecolor{4}{rgb}{0.990,0.796,0.494}
\definecolor{3}{rgb}{1.00,0.92156,0.5176}
\definecolor{2}{rgb}{0.6941,0.8314,0.498}
\definecolor{1}{rgb}{0.388,0.745,0.4823}
\newcommand{\first}[1]{%
  \begingroup
  \setlength{\fboxsep}{2pt}%  
  \colorbox{1}{#1}%
  \endgroup
}
\newcommand{\second}[1]{%
  \begingroup
  \setlength{\fboxsep}{2pt}%  
  \colorbox{2}{#1}%
  \endgroup
}
\newcommand{\third}[1]{%
  \begingroup
  \setlength{\fboxsep}{2pt}%  
  \colorbox{3}{#1}%
  \endgroup
}
\newcommand{\fourth}[1]{%
  \begingroup
  \setlength{\fboxsep}{2pt}%  
  \colorbox{4}{#1}%
  \endgroup
}
\newcommand{\fifth}[1]{%
  \begingroup
  \setlength{\fboxsep}{2pt}%  
  \colorbox{5}{#1}%
  \endgroup
}
\newcommand{\sixth}[1]{%
  \begingroup
  \setlength{\fboxsep}{2pt}%  
  \colorbox{6}{#1}%
  \endgroup
}
\newcommand{\seventh}[1]{%
  \begingroup
  \setlength{\fboxsep}{2pt}%  
  \colorbox{7}{#1}%
  \endgroup
}
\newcommand{\eigth}[1]{%
  \begingroup
  \setlength{\fboxsep}{2pt}%  
  \colorbox{8}{#1}%
  \endgroup
}

\newcommand{\etal}{\emph{et al}.}

\begin{document}
% \linenumbers
\pagestyle{headings}
\mainmatter
\def\ECCVSubNumber{5227}  % Insert your submission number here

\title{GroSS: Group-Size Series Decomposition for Grouped Architecture Search} % Replace with your title

\index{Howard-Jenkins, Henry}
\index{Prisacariu, Victor Adrian}

\begin{comment}
\titlerunning{ECCV-20 submission ID \ECCVSubNumber} 
\authorrunning{ECCV-20 submission ID \ECCVSubNumber} 
\author{Anonymous ECCV submission}
\institute{Paper ID \ECCVSubNumber}
\end{comment}
%******************

% \begin{comment}
\titlerunning{GroSS Decomposition}
\author{Henry Howard-Jenkins \and
Yiwen Li \and
Victor Adrian Prisacariu}
\authorrunning{H. Howard-Jenkins et al.}
\institute{Active Vision Laboratory, University of Oxford, UK \\ \email{\{henryhj, kate, victor\}@robots.ox.ac.uk}}
% \end{comment}
%******************
\maketitle

\begin{abstract}
We present a novel approach which is able to explore the configuration of grouped convolutions within neural networks. Group-size Series (GroSS) decomposition is a mathematical formulation of tensor factorisation into a series of approximations of increasing rank terms. GroSS allows for dynamic and differentiable selection of factorisation rank, which is analogous to a grouped convolution. Therefore, to the best of our knowledge, GroSS is the first method to enable simultaneous training of differing numbers of groups within a single layer, as well as all possible combinations between layers. In doing so, GroSS is able to train an entire grouped convolution architecture search-space concurrently. We demonstrate this through architecture searches with performance objectives on multiple datasets and networks. GroSS enables more effective and efficient search for grouped convolutional architectures.
\end{abstract}

\section{Introduction}

In recent years, there has been a flurry of deep neural networks (DNNs) producing remarkable results on a broad variety of tasks.
% Generally, these methods have usually involved careful network design, often relying on domain knowledge to design a structure which can encapsulate the task at hand. 
In particular, grouped convolution has become a widely used tool in some prevalent networks. ResNeXt~\cite{xie2017aggregated} used grouped convolution for improved accuracy over the analogous ResNets~\cite{he2016deep}. On the other hand, Xception~\cite{chollet2017xception}, MobileNet~\cite{howard2017mobilenets}, and various others~\cite{zhang2018shufflenet,sandler2018mobilenetv2} have used depthwise convolutions, which are the special case of grouped convolutions where the number of groups is equal to the number of in channels, in a for extremely low-cost inference. With these architectures, grouped convolution has proven to be a valuable design tool for high-performance and low-cost design alike. But, its application to these contrasting performance profiles has so far, to the best of our knowledge, remained relatively unexplored.

Finding a heuristic or intuition for how combinations of grouped convolutions with varying numbers of groups interact within a network is challenging. Grouped convolution, therefore, is presents itself as an ideal candidate for Neural Architecture Search (NAS), which has provided an alternative to hand designed networks. NAS allows for the search and even direct optimisation of the network's structure. But, the search space for architectures is often vast, with potentially limitless design choices. Furthermore, each configuration must undergo some training or fine-tuning for its efficacy to be determined. This has led to the development of methods which lump multiple design parameters together, which reduce the search space in a principled manner~\cite{tan2019efficientnet}, as well as creating the need for sophisticated search algorithms~\cite{liu2018darts,wu2019fbnet}, which can more quickly converge to an improved design. Both techniques reduce the number of search iterations and ultimately reduce the number of required training/fine-tuning stages.

In this work, however, we do not wish to make assumptions about the grouped convolution manifold. We achieve this with the introduction of a Group-size Series (GroSS) decomposition. GroSS allows us to train the entire search space of architectures \textit{simultaneously}. In doing so, we shift the expense of architecture search with respect to groups away from decomposition and training, and towards cheaper test-time sampling. This allows for the exploration of possible configurations, while significantly reducing the need for imparting bias on the group design hyperparameter selection.

The contributions of this paper can be summarised as follows:
\begin{enumerate}
    \item We present GroSS decomposition -- a novel formulation of tensor decomposition as a series of rank approximations. This provides a mathematical basis for grouped convolution as a series of increasing rank terms.
    \item GroSS provides the apparatus for differentiably switching between grouped convolution ranks. Therefore, to the best of our knowledge, it is the first simultaneous training of differing numbers of groups within a single layer, as well as the all possible configurations between layers. This makes feasible, for the first time, a search for rank selection for network compression.
    \item We explore this concurrently-trained architecture space in the context of network acceleration. We factorise a small network, as well VGG-16 and ResNet-18, and propose exhaustive and breadth-first searches on CIFAR-10 and ImageNet. We demonstrate the efficacy of the GroSS for rank selection search over a more conventional approach of partial training schedules.
\end{enumerate}

\section{Related Work}

Grouped convolution has had a wide impact on neural network architectures, particularly due to its efficiency. It was first introduced in AlexNet~\cite{krizhevsky2012imagenet} as an aid for the single network to be trained over multiple GPUs. Since then, it has had a wide impact on DNN architecture design. Deep Roots~\cite{ioannou2017deep} was the first to introduce group convolution for efficiency, while ResNeXt~\cite{xie2017aggregated} used grouped convolutions synonymously with concept of \textit{cardinality}, ultimately exploiting the efficiency of grouped convolutions for high-accuracy network design. The reduced complexity of grouped convolution allowed for ResNeXt to incorporate deeper layers within the ResNet-analogous residual blocks~\cite{he2016deep}. In all, this allowed higher accuracy with a similar inference cost as an equivalent ResNet. The efficiency of grouped convolution has also led to several low-cost network designs. Sifre~\cite{sifre2014rigid} first introduced depthwise separable convolutions, which were later utilised by Xception~\cite{chollet2017xception}. MobileNet~\cite{howard2017mobilenets} utilised a ResNet-like bottleneck design with depthwise convolutions for an extremely efficient network with mobile applications in mind. ShuffleNet~\cite{zhang2018shufflenet} was also based on a depthwise bottleneck, however, pointwise layers were also made grouped convolutions.

Previous works~\cite{jaderberg2014speeding,denton2014exploiting,lebedev2014speeding,vanhoucke2011improving} have applied low-rank approximation of convolution for network compression and acceleration. Block Term Decomposition (BTD)~\cite{de2008decompositions} has recently been applied to the task of network factorisation~\cite{chen2018sharing}, where it was shown that the BTD factorisation of a convolutional weight was equivalent to a grouped convolution within a bottleneck architecture. Wang~\etal~\cite{wang2018deepsearch} applied this equivalency for network acceleration. Since decomposition is costly, these methods have relied on heuristics and intuition to set hyperparameters such as the rank of successive layers within the decomposition. In this paper, we present a method for decomposition which allows for exploration of the decomposition hyperparameters and all the combinations.

Existing architecture search methods have overwhelmingly favoured reinforcement learning. Examples of this include, but are not limited to, NASNet~\cite{zoph2018learning}, MNasNet~\cite{tan2019mnasnet}, ReLeq-Net~\cite{elthakeb2018releq}. In broad terms, these methods all set a baseline structure, which is manipulated by a separate controller. The controller optimises the structure through and objective based on network performance. There has also been work in differentiable architecture search~\cite{wu2019fbnet,liu2018darts} which makes the network architecture manipulations themselves differentiable. In addition, work such as~\cite{tan2019efficientnet} aims to limit the network scaling within a performance envelope to a single parameter.

These methods all have a commonality: the cost of re-training or fine-tuning at each stage motivates the recovery of the optimal architecture in as few training steps as possible, whether this is achieved through a trained controller, direct optimisation or significantly reducing the search space. In this work, however, GroSS allows efficient weight-sharing between varying grouped architectures, thus enabling them to be trained at once. This is similar to the task of one-shot architecture search. SMASH~\cite{brock2017smash} use a hypernetwork to predict weights for each architecture. The work of Li~\etal~\cite{li2019random} bares most resemblance to this paper, where randomly sampled architectures are used to train shared-weights.

\section{Method}
In this section, we will first introduce Block Term Decomposition (BTD) and detail how its factorisation can be applied to a convolutional layer. After that, we will introduce GroSS decomposition, where we formulate a unification of a series of ranked decompositions so that they can dynamically and differentially be combined. We detail the training strategy for training the whole series at once. Finally, we detail the methodology of our exhaustive and breadth-first search.

\subsection{General Block Term Decomposition}
Block Term Decomposition (BTD)~\cite{de2008decompositions} aims to factorise a tensor into the sum of multiple low rank-Tuckers~\cite{tucker1966some}. That is, given an $N^{\text{th}}$ order tensor $\tX \in \sR^{d_1 \times d_2 \times ... \times d_N}$, BTD factorises $\tX$ into the sum of $R$ terms with rank $(d'_1, d'_2, ..., d'_N)$:
\begin{equation}
\begin{split}
    \tX = \sum^R_{r=1}\tG_r \times_1 \tA_r^{(1)} \times_2 \tA_r^{(2)} \times_3 ... \times_N \tA_r^{(N)} \\
\text{where}
\begin{cases}
     & \tG \in \sR^{d'_1 \times d'_2 \times ... \times d'_N} \\
     & \tA_r^{(n)} \in \sR^{d_n \times d'_n}, n \in \{1, ..., N\}
\end{cases}
\end{split}
\end{equation}

In the above, $\tG$ is known as the \emph{core} tensor and we will refer to $\tA$ as \emph{factors} matrices. We use the usual notation $\times_n$ to represent the \textit{mode-n} product~\cite{de2008decompositions}.

\subsection{Converting a Single Convolution to a Bottleneck Using BTD}

Here, we can restrict discussion from a general, N-mode, tensor to the 4-mode weights of a 2D convolution as follows: $\tX \in \sR^{t \times u \times v \times w}$, where $t$ and $u$ represent the number of input and output channels, and $v$ and $w$ the spatial size of the filter kernel. Typically the spatial extent of each filter is small and thus we only factorise $t$ and $u$. To eliminate superscripts, we define $\tB = \tA^{(1)}$ and $\tC = \tA^{(2)}$. Therefore, the BTD for convolutional weights is expressed as follows:
\begin{equation}
\begin{split}
    \tX = \sum^R_{r=1}\tG_r \times_1 \tB_r \times_2 \tC_r \\
        \text{where}
        \begin{cases}
             & \tG \in \sR^{t' \times u' \times v \times w} \\
             & \tB \in \sR^{t \times t'} \\
             & \tC \in \sR^{u \times u'} \\
        \end{cases}
\end{split}
\end{equation}

\begin{figure}[b]
    \centering
    \includegraphics[width=0.65\textwidth]{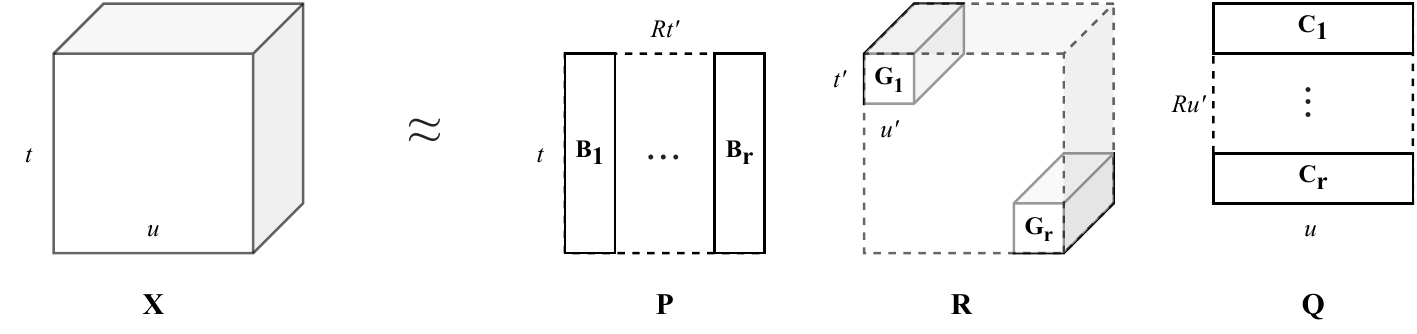}
    \caption{Formation of bottleneck layers $\tP$, $\tR$ and $\tQ$ from BTD cores and factors.}
    \label{fig:formation}
\end{figure}

This factorisation of the convolutional weights into $R$ groups forms a three-layer bottleneck-style structure~\cite{chen2018sharing}: a pointwise ($1\times1$) convolution $\tP \in \sR^{t \times (Rt') \times 1 \times 1}$, formed from factor $\tB$; followed by a grouped convolution $\tR \in \sR^{t' \times (Ru') \times v \times w}$, formed from core $\tG$ and with $R$ groups; and finally another pointwise convolution $\tQ \in \sR^{(Ru') \times u \times 1 \times 1}$, formed from factor $\tC$.
With careful selection of the BTD parameters, the bottleneck approximation can be applied to any standard convolutional layer. This is visualised in Figure~\ref{fig:formation}.
% \begin{equation}
%     \begin{cases}
%     R & = \text{Number of groups in the grouped convolution} \\
%     t & = \text{Number of input channels} \\
%     u & = \text{Number of output channels} \\
%     t' = u' & = \dfrac{\text{Bottleneck width}}{R} = \text{Group-size}
%     \end{cases}
% \end{equation}

In Table~\ref{tab:conv_setup}, we detail how the dimensions of the bottleneck architecture are determined from its corresponding convolutional layer, and indicate how properties such as stride, padding and bias are applied within the bottleneck for equivalency with the original layer. It is worth noting that we often refer to the quantities $t'$ or $u'$ as the group-size; this quantity determines the number of channels in each group and is equivalent to the rank of the decomposition.

\begin{table}[t]
    \centering
    \caption{Convolution to grouped bottleneck. The table states how the convolutional parameters are used in the equivalent bottleneck}
    \begin{tabular}{cccccccc}
    \toprule
        & Filter Size & $C_{in}$ & $C_{out}$ & Groups & Bias & Stride & Padding         \\ \midrule
        Convolution & $v \times w$  & $t$ & $u$ & 1     & $B$   & $S$   & $P$   \\ \addlinespace[0.25em]
                    & $1 \times 1$  & $t$ & $Rt'$ & 1     & -     &  1  &  0  \\
        Bottleneck  & $v \times w$  & $Rt'$ & $Ru'$ & $R$   & -     &  $S$    & $P$     \\
                    & $1 \times 1$  & $Ru'$ & $u$ & 1     & $B$   &  1    & 0     \\
    \bottomrule
    \end{tabular}
    \label{tab:conv_setup}
\end{table}

\subsection{Group-size Series Decomposition}
Group-size Series (GroSS) decomposition unifies multiple ranks of BTD factorisations. This is achieved by defining each successive factorisation relative to the lower order ranks. Thus we ensure that higher rank decompositions only contain information that was missed by the lower order approximations. Therefore the $i^{\text{th}}$ approximation of $\tX$ is given as follows:
\begin{equation}
\begin{split}
    \tX = \sum^{R_i}_{r=1}[(\tens{g}_r)_i + (\tG'_r)_{i-1}] 
        \times_1 [(\tens{b}_r)_i + (\tB'_r)_{i-1}]
        \times_2 [(\tens{c}_r)_i + (\tC'_r)_{i-1}] \\
    \text{where}
    \begin{cases}
         & (\tens{g}_r)_i,\ (\tG'_r)_{i-1} \in \sR^{t'_i \times u'_i \times v \times w} \\
         & (\tens{b}_r)_i,\ (\tB'_r)_{i-1} \in \sR^{t \times t'_i} \\
         & (\tens{c}_r)_i,\ (\tC'_r)_{i-1} \in \sR^{u \times u'_i} \\
    \end{cases}
\end{split}
\end{equation}

Omitting $r$ from the notation, $\tens{g}_i$, $\tens{b}_i$ and $\tens{c}_i$ represent the additional information captured between the $(i-1)^{\text{th}}$ and $i^{\text{th}}$ rank of approximation, and $\tG_{(i-1)}'$, $\tB_{(i-1)}'$ and $\tC_{(i-1)}'$ to represent total approximation from lower rank approximations in the form of cores and factors. However, both the core and factors must be recomputed so that the dimensions match the ranks required $R_i$, which is not a trivial manipulation.

\begin{figure}
    \centering
    \includegraphics[width=0.60\textwidth]{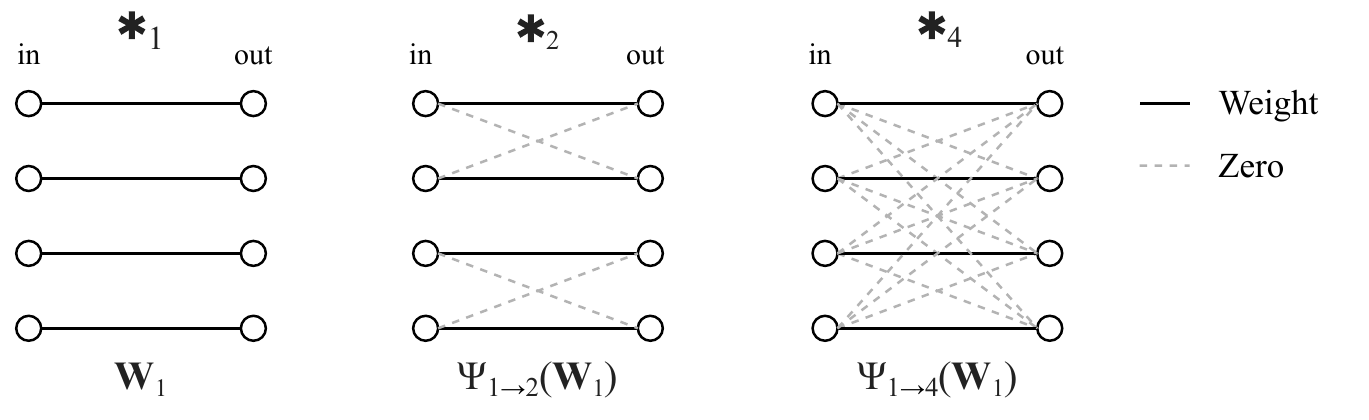}
    \caption{Visualisation of a depthwise weight, $\tW_1$, expanded for convolutions with groups of size 2 and 4.}
    \label{fig:expansion}
\end{figure}

Instead, we introduce a function,  $\Psi_{g \rightarrow h} ()$, which allows the weights of a grouped convolution to be ``expanded". The expanded weight from a convolution with group-size $g$ can be used in a convolution with group-size $h$, where $h>g$, giving identical outputs:
\begin{equation}
    \tW_g *_g \mF  \equiv \Psi_{g \rightarrow h}(\tW_g) *_h \mF \\
\end{equation}

\noindent where $\tW_g$ is the weight for a grouped convolution, $*_g$ refers to convolution with group-size $g$, and $\mF$ is the feature map to which the convolution is applied. We provide an example visualisation of $\Psi()$ in Figure~\ref{fig:expansion}. This allows us to conveniently reformulate the GroSS decomposition in terms of the successive convolutional weights obtained from BTD, rather than within the cores and factors directly. More specifically, we define the bottleneck weights for the $N^{\text{th}}$ order GroSS decomposition with group-sizes, $S = \{s_1, ..., s_N\}$, as follows:
\begin{equation}
\begin{split}
    \tR_{N} = \Psi_{s_1 \rightarrow s_N}(\tR_1) + \sum^N_{i=2}\Psi_{s_i \rightarrow s_N}(\tens{r}_{i}) \\
    \tP_{N} = \tP_{1} + \sum^N_{i=2}\tens{p}_{i}, \quad
    \tQ_{N} = \tQ_{1} + \sum^N_{i=2}\tens{q}_{i}
\end{split}
\label{eq:convexp}
\end{equation}

$\tR_{1}$, $\tP_{1}$ and $\tQ_{1}$ represent the weights obtained from the lowest rank decomposition present in the series. $\tens{r}_{i}$, $\tens{p}_{i}$ and $\tens{q}_{i}$ represent the additional information that the $i^\text{th}$ rank decomposition contribute to the bottleneck approximation:
\begin{equation}
    \tens{p}_{i} = \tP_{i} - \tP_{(i-1)}, \quad
    \tens{r}_{i} = \tR_{i} - \Psi_{s_{(i-1)} \rightarrow s_i}(\tR_{(i-1)}), \quad
    \tens{q}_{i} = \tQ_{i} - \tQ_{(i-1)}.
\end{equation}
This formulation involving only manipulation of the convolutional weights is exactly equivalent to forming the bottleneck components $\tens{r}_{i}$, $\tens{p}_{i}$ and $\tens{q}_{i}$ from $\tens{g}_{i}$, $\tens{b}_{i}$ and $\tens{c}_{i}$, as in the general BTD to bottleneck case.

Further, the grouped convolution weight expansion, $\Psi()$, enables us to dynamically, and differentiably, change the group-size of a convolution. In itself, this is not particularly useful: a convolution with a larger group-size is requires more operations and more memory, while yielding identical outputs. However, it allows for direct interaction between differently ranked network decomposition and, therefore, the representation of one rank by the combination of lower ranks. Thus, GroSS treats the decomposition of the original convolution as the sum of successive order approximations, with each order contributing additional representational power.

\subsubsection{Training GroSS Simultaneously}
The expression of a group-size $s_i$ decomposition as the combination of lower rank decompositions is useful because it enables the group-size to be dynamically changed during training. The expansion and summation of convolutional weights is differentiable and so training at a high rank, also optimises the lower rank approximations simultaneously. To the best of our knowledge GroSS is the first method that allows weight-sharing between, and training of, convolutions with varying numbers of groups.

We leverage the series form of the factorisation during training, by randomly sampling a group-size for each decomposed layer at each iteration. We sample a group-size $s_i$ for each decomposed layer uniformly. Through uniform sampling, we are able to train each network configuration equally.

\subsection{Search}
\label{app:bsearch}
The objective for all the searches performed within in this paper is: given a base configuration, we aim to find an alternative configuration which is more accurate, but offers the same or cheaper inference. We implement two forms of search to achieve this leveraging GroSS decomposition: exhaustive and breadth-first. 

\subsubsection{Exhaustive.} Within the exhaustive search, all possible configurations are evaluated. In this search, we simply filter any configuration with multiply accumulates (MACs) above the respective base configuration. After filtering, we can select the highest accuracy remaining.

\subsubsection{Breadth-first Search.} Where an exhaustive search is not feasible due to the sheer number of possible configurations, we use a greedy breadth-first search. We first randomly select a configuration which requires fewer operations for inference than the base configuration. We evaluate all neighbouring configurations---those which only require one layer to have it's group-size changed---of the currently selected configuration. We select the neighbour with the highest accuracy that does not exceed the number of MACs as the base configuration for the next step. We repeat this step for a maximum of 25 times, or until there are no more accurate neighbours not exceeding the cost of the base configuration.

This is repeated 20 times. The most accurate configuration from all of the 20 runs is considered the result of the search. Since the search for a base configuration with fewer MACs is contained within the search-space with a higher limit, we perform them incrementally. This results in the same search process, but the first 10 runs are initialised using the top-10 highest accuracy results from the smaller search. We found that this generally led to faster stopping.

\section{Application of GroSS}
In this section, we explain how we apply GroSS to a several models across datasets. We first detail the dataset on which evaluation is conducted. Next, we describe the network architecture on which perform GroSS decomposition. Finally, we list the procedure for the decomposition and fine-tuning. 

\subsection{Datasets}
We perform our experimental evaluation on CIFAR-10~\cite{krizhevsky2014cifar} and ImageNet~\cite{krizhevsky2012imagenet}. CIFAR-10 is a dataset consisting of 10 classes. The size of each image is $32 \times 32$. In total there are 60,000 images, which are split into 50,000 train images and 10,000 testing images. We further divide the training set into a training and validation splits with 40,000 and 10,000 images, respectively. ImageNet consists of 1000 classes, with 1.2 million training images and 50,000 validation images. Since the test annotations are not available, we report our accuracy on the validation set.

\subsection{Models}
In this paper, we perform on three general network architectures: a custom 4-layer network, VGG-16~\cite{simonyan2014very}, and ResNet-18~\cite{he2016deep}. Here, we provide an overview of the network definitions, with more details in Appendices~\ref{app:network} and~\ref{app:training}.

Our 4-layer network has four convolutional layers, with output channel dimensions of 32, 32, 64 and 64, followed by two fully-connected layers of size 256 and 10. In our ImageNet experiments, we use a standard VGG-16 and ResNet-18, identical to those in~\cite{simonyan2014very} and~\cite{he2016deep}, respectively. However, we make some changes to the fully connected structure in VGG-16 for training and inference on CIFAR-10. The convolutional layers instead followed by a $2 \times 2$ max-pooling and two fully-connected layers of size 512 and 10, respectively. A ReLU layer and dropout with probability of 0.5 is applied between the fully-connected layers.

\subsection{Decomposition}
We perform GroSS decomposition on our small 4-layer network, as well as VGG-16~\cite{simonyan2014very}. In each case we decompose all convolutional layers in the network aside from the first. Unless otherwise stated, we set the bottleneck width equal to the number of input channels. For the 4-layer network, group-sizes are set to all powers of 2 which do not exceed the bottleneck width for that respective layer. This leads to a total of 252 configurations represented by our decomposition. We decompose each layer in VGG-16 and ResNet-18 into 4 group-sizes: (1, 4, 16, 32). This leads to a total of $4^{12}$ and $4^{16}$ configurations represented by the decomposed VGG-16 and ResNet-18, respectively.

Our formulation of GroSS decomposition as a series of convolutional weight differences (expanded weights in the case of the grouped convolution), as detailed by Equation~\ref{eq:convexp} means that we are able to use an off-the-shelf BTD framework~\cite{kossaifi2019tensorly}. For each group-size, we set the stopping criteria for BTD identically: when the decrease in approximation error between steps is below $1 \times 10^{-6}$ for the 4-layer network and $1 \times 10^{-5}$ for VGG-16, or $5 \times 10^5$ steps have elapsed. We define approximation error as the Frobenius norm between the original tensor and the product of the BTD cores and factors divided by the Frobenius norm of the original tensor. For the 4-layer network, we perform this decomposition 5 times.

\subsection{Fine-tuning}
\label{sec:cifarft} 
\subsubsection{CIFAR-10.} After we have performed GroSS decomposition on the network, we then fine-tune on the classification task. For the 4-layer network, we tune for 150 epochs with a batch-size of 256, an initial learning rate of 0.0001 and momentum 0.9. We decay the learning rate by a factor of 0.1 after both 80 and 120 epochs. For VGG-16, we fine-tune with the same SGD parameters and batch-size, however we train for 200 epochs, and decay the learning rate after 100 and 150 epochs. All network parameters are frozen aside from the GroSS decomposition weights. During training, there is a 0.5 probability of horizontal flipping, zero-padding of size 2 is applied around all borders and a random $32 \times 32$ crop is taken from the resulting image.

\subsubsection{ImageNet.} We decompose VGG-16 and ResNet-18 before funetuning on ImageNet. For VGG-16, we train using SGD for a total of 4 epochs with a batch-size of 128, leading to approximately $10^4$ iterations. The initial learning rate is set to $10^{-5}$, which is decayed by a factor of 0.1 after 2 epochs. Momentum is set to 0.9. Again, all the network parameters are frozen, aside from the decomposition weights. For ResNet-18, we train for 8 epochs in total with a batch size of 512. The initial learning rate is set to $5\times10^{-5}$, with decay every 2 epochs  The images are resized so that the smallest side is of size 256. During training, the resized images are flipped horizontally with a probability of 0.5 and a random $224 \times 224$ crop is taken. During testing, we simply take a centre crop from the resized image, hence evaluating 1-crop accuracy.

\subsubsection{Individual Configurations.}
\label{sec:indconf}
For the decomposition in the conventional manner, \textit{i.e.} a singular group-size configuration, we decompose using exactly the same routine as with GroSS. However, the fine-tuning schedules are slightly modified. On CIFAR-10, we reduce the schedule for our 4-layer network and our CIFAR VGG-16 to 100 epochs. The initial learning rate is increased to 0.001, and decayed at 80 epochs. On Imagenet, we do not freeze the non-decomposed layers. The VGG-16 configurations have a schedule of 6 epochs, with learning rate decay occurring after every 2 epochs. The initial learning rate is kept at $10^{-4}$. For ResNet-18 configurations, we increase batch size to 512 and again unfreeze all layers. We run for a total of 12 epochs, with initial learning rate $10^{-3}$ and decay after 8 and 10 epochs. Due to this being the conventional BTD factoristation strategy, we often refer to this as the \textit{true} accuracy of a configuration.

\begin{figure}
\begin{center}
\includegraphics[width=0.70\textwidth]{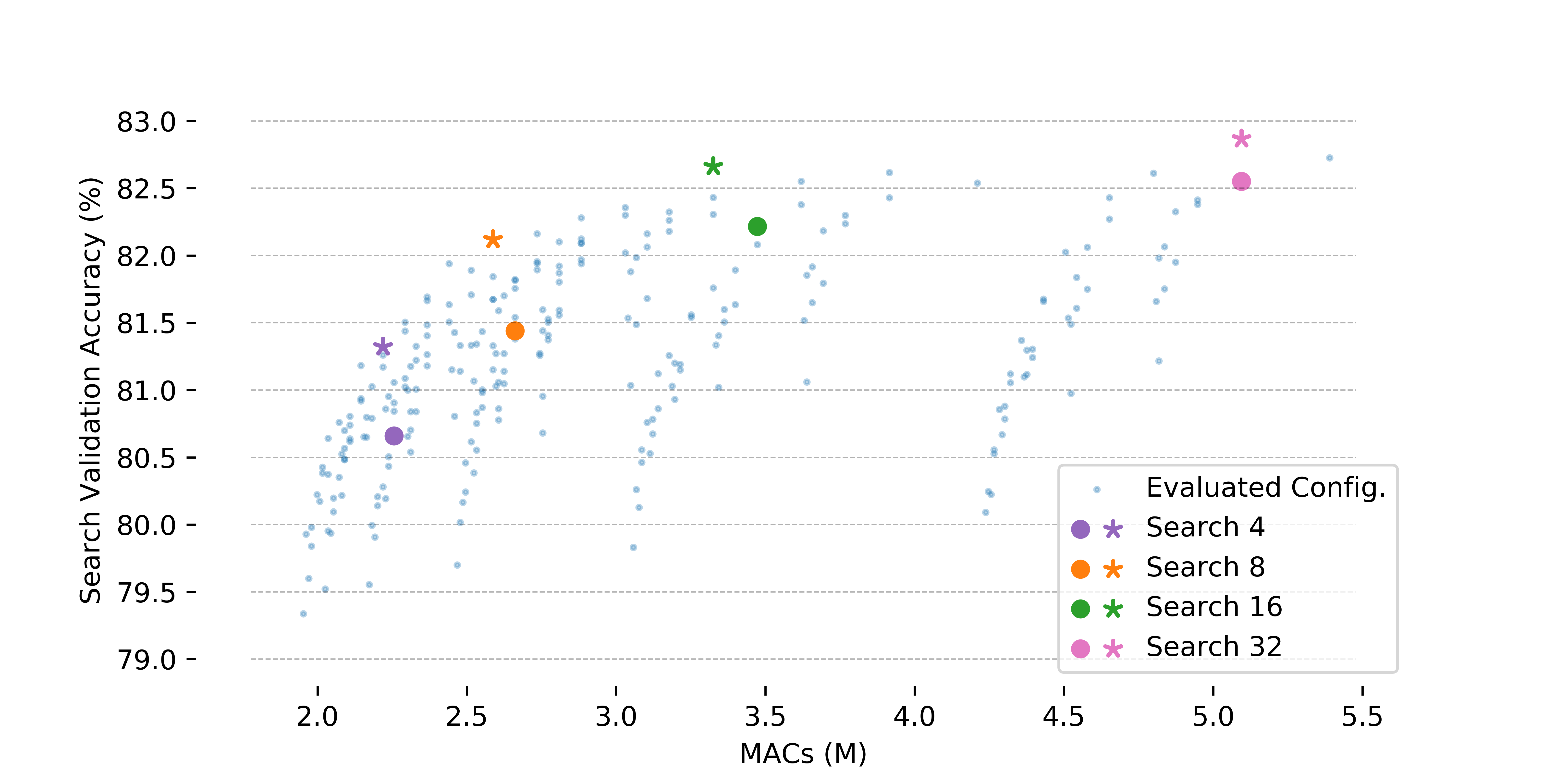}
\end{center}
\caption{Exhaustive search for the 4-layer network on CIFAR-10. Each search is colour coded. The circles and stars mark the performance of the baseline and found configuration, respectively.}
\label{fig:4layer_search}
\end{figure}

\section{Results}

In this section, we demonstrate the effectiveness of GroSS. First, we explore group-size selection for network acceleration through search on our GroSS decomposition. Secondly, we justify the design of GroSS over a more simply using partial training schedules for group configurations.

\subsection{Group-size Search}

Here, we evaluate the performance of our search. We split our results by dataset, with our CIFAR-10 results being followed by our results on ImageNet.

For each search, we report the change in accuracy of the found configuration over the baseline, as well as the percentage reduction in MACs compared to the baseline. We are primarily concerned with exploring the impact of the number of groups on the performance of the network, rather than other design parameters such as the bottleneck dimensions. In our experimental setup, the inference cost between configurations varies only in the number of groups in each bottleneck. However, there is significant overhead from other layers in the network that remains constant between configurations. Therefore, we report the reduction in total MACs, as well as the contribution from the group layers alone. This provides greater insight into grouped architecture design and the performance of the searches using GroSS.

\begin{table}
    \centering
    \caption{Exhaustive search for our 4-layer network on CIFAR-10}
    \begin{tabular}{ccccccc}
    \toprule
        \textbf{4-Layer Network}& & \multicolumn{2}{c}{\textbf{Accuracy}} & & \multicolumn{2}{c}{\textbf{$\Delta$MACs}} \\
        \textbf{Configuration} & \textbf{MACs} & GroSS & True & $\Delta$Acc. & Total & G.Conv \\ \midrule
        
        Full & 5.13M & - & 83.99 (0.53) & - & - & -\\ \addlinespace[0.25em]
        
        \textit{Baseline: \sixth{32}\sixth{32}\sixth{32}} & \textbf{\textit{5.09M}} & \textit{82.55 (0.07)} & \textit{83.70 (0.05)} & - & - & -\\
        \sixth{32}\fifth{16}\seventh{64} & \textbf{5.09M} & \textbf{82.87 (0.14)} & \textbf{84.05 (0.07)} & $\mathbf{\uparrow 0.35}$ & 0.00\% & 0.00\% \\ \addlinespace[0.25em]
        
        \textit{Baseline: \fifth{16}\fifth{16}\fifth{16}} & \textit{3.47M} & \textit{82.22 (0.10)} & \textit{82.94 (0.06)} & - & - & -\\
        VBMF \cite{nakajima2013global}: \fifth{16}\fourth{8}\fifth{16}& \textbf{3.33M} & 81.76 (0.10) & 82.83 (0.10) & $\downarrow0.11$& $\mathbf{\downarrow4.25\%}$  & $\mathbf{\downarrow9.09\%}$ \\
        \fourth{8}\fifth{16}\seventh{64} & \textbf{3.33M} & \textbf{82.66 (0.11)} & \textbf{83.88 (0.10)} & $\mathbf{\uparrow 0.94}$ & $\mathbf{\downarrow4.25\%}$ &  $\mathbf{\downarrow9.09\%}$ \\ \addlinespace[0.25em]
        
        \textit{Baseline: \fourth{8}\fourth{8}\fourth{8}} & \textit{2.66M} & \textit{81.44 (0.16)} & \textit{82.86 (0.10)} & - & - & -\\
        \second{2}\fifth{16}\sixth{32} & \textbf{2.59M} & \textbf{82.12 (0.16)} & \textbf{83.50 (0.07)} & $\mathbf{\uparrow 0.64}$ & $\mathbf{\downarrow2.77\%}$ & $\mathbf{\downarrow9.09\%}$ \\ \addlinespace[0.25em]
        
        \textit{Baseline: \third{4}\third{4}\third{4}} & \textit{2.26M} & \textit{80.66 (0.13)} & \textit{82.37 (0.11)} & - & - & -\\
        \first{1}\fourth{8}\fifth{16} & \textbf{2.22M} & \textbf{81.32 (0.16)} & \textbf{82.45 (0.15) } & $\mathbf{\uparrow 0.08}$ & $\mathbf{\downarrow1.63 \%}$ & $\mathbf{\downarrow9.09\%}$ \\ \addlinespace[0.25em]
        
        \textit{Depthwise: \first{1}\first{1}\first{1}} & \textit{1.95M} & \textit{79.34 (0.14)} & \textit{81.70 (0.32)} & - & - & -\\
    \bottomrule
    \end{tabular}
    \label{tab:search}
\end{table}

\subsubsection{CIFAR-10.} The results of the exhaustive search on the 4-layer network are shown in Table~\ref{tab:search}, where the decomposition and tune is performed 5 times for each configuration and the mean and standard deviation are reported decompositions with uniform rank values across layers (4, 8, 16, and 32) are chosen as the baseline configurations for the search, such that we perform search across the range of possible configurations. For each baseline configuration we are able to find an alternative that is more accurate whilst requiring fewer operations. The results of the search are also visualised in Figure~\ref{fig:4layer_search}.

In the case of our CIFAR-10 VGG-16 network, the $4^{12}$ configurations produced by our GroSS decomposition are too many to feasibly enable exhaustive evaluation. We, therefore, perform a breadth-first search. The full details of how this search is performed are described in Section~\ref{app:bsearch}. Results for this search on VGG-16 are shown in Table~\ref{tab:vgg_search}.

\begin{table}[t]
    \centering
    \caption{Breadth-first search on our VGG-16 network on CIFAR-10}
    \resizebox{\columnwidth}{!}{%
    \begin{tabular}{ccccccc}
    \toprule
        \textbf{VGG-16 (CIFAR)}& & \multicolumn{2}{c}{\textbf{Accuracy}} & & \multicolumn{2}{c}{\textbf{$\Delta$MACs}} \\
        \textbf{Configuration} & \textbf{MACs} & GroSS & True & $\Delta$Acc. & Total & G.Conv \\ \midrule
        Full & 314M & - & 91.52 & - & - & -\\ \addlinespace[0.25em]
        
        \textit{\sixth{32}\sixth{32}\sixth{32}\sixth{32}\sixth{32}\sixth{32}\sixth{32}\sixth{32}\sixth{32}\sixth{32}\sixth{32}\sixth{32}} & \textit{121M} & \textit{90.97} & \textbf{\textit{91.57}} & - & - & -\\
        VBMF \cite{nakajima2013global} & 118M & 90.97 & 91.31 & $\downarrow0.26$ & $\downarrow2.68\%$ &  $\mathbf{\downarrow6.18 \%}$ \\
        \sixth{32}\third{4}\fifth{16}\sixth{32}\fifth{16}\fifth{16}\sixth{32}\fifth{16}\third{4}\fifth{16}\sixth{32}\first{1} & \textbf{103M} & \textbf{91.31} & 91.41 & $\downarrow0.16$ & $\mathbf{\downarrow14.6 \%}$ &  $\mathbf{\downarrow33.7 \%}$ \\ \addlinespace[0.25em]
    
        \textit{\fifth{16}\fifth{16}\fifth{16}\fifth{16}\fifth{16}\fifth{16}\fifth{16}\fifth{16}\fifth{16}\fifth{16}\fifth{16}\fifth{16}} & \textit{94.6M} & \textit{91.13} & \textit{91.19} & - & - & -\\
        \third{4}\third{4}\fifth{16}\sixth{32}\fifth{16}\fifth{16}\first{1}\sixth{32}\third{4}\sixth{32}\fifth{16}\first{1} & \textbf{86.7M} & \textbf{91.28} & \textbf{91.31} & $\mathbf{\uparrow 0.12}$ & $\mathbf{\downarrow8.36\%}$ &  $\mathbf{\downarrow30.2 \%}$ \\ \addlinespace[0.25em]
    
        \textit{\third{4}\third{4}\third{4}\third{4}\third{4}\third{4}\third{4}\third{4}\third{4}\third{4}\third{4}\third{4}} & \textit{74.9M} & \textit{90.43} & \textit{90.90} & - & - & -\\
        \first{1}\first{1}\first{1}\fifth{16}\fifth{16}\third{4}\first{1}\first{1}\third{4}\third{4}\first{1}\third{4} & \textbf{74.1M}& \textbf{90.97} & \textbf{91.14} & $\mathbf{\uparrow 0.24}$ & $\mathbf{\downarrow1.11 \%}$ &  $\mathbf{\downarrow12.6 \%}$ \\ \addlinespace[0.25em]
        
        \textit{\first{1}\first{1}\first{1}\first{1}\first{1}\first{1}\first{1}\first{1}\first{1}\first{1}\first{1}\first{1}} & \textit{70.0M} & \textit{90.24} & \textit{90.66} & - & - & -\\
    
    \bottomrule
    \end{tabular}
    }
    \label{tab:vgg_search}
\end{table}

For the searches on the 4, 8 and 16 baselines, we are able to find configurations which meet the objective. However, in the case of the search below the 32 baseline, the found configuration's true accuracy is less than that of the baseline. We speculate that this is because 32 is the maximum rank in the decomposition. Therefore, the rank of each layer can never be increased above that of the baseline. This means that configurations have less room to manoeuvre in targeting more heavy-duty layers at key stages of the network.

We also include a configuration found through Variational Bayesian Matrix Factorisation (VBMF)~\cite{nakajima2013global}, which is used for one-shot rank selection in \cite{kim2015compression}. For both networks, we were able to find more accurate configurations which require fewer or the same number of operations than the VBMF rank selection. In fact, Kim~\etal~\cite{kim2015compression} note that, although they achieve good network compression results with the result of VBMF, they had not investigated whether this method of rank selection was optimal. The results in Table~\ref{tab:search} demonstrate that VBMF is not optimal in this case, and GroSS is an effective tool to determine this.

\subsubsection{ImageNet.} 
\begin{figure}[b]
\begin{center}
%\framebox[4.0in]{$\;$}
\subfloat{\includegraphics[width=0.45\textwidth]{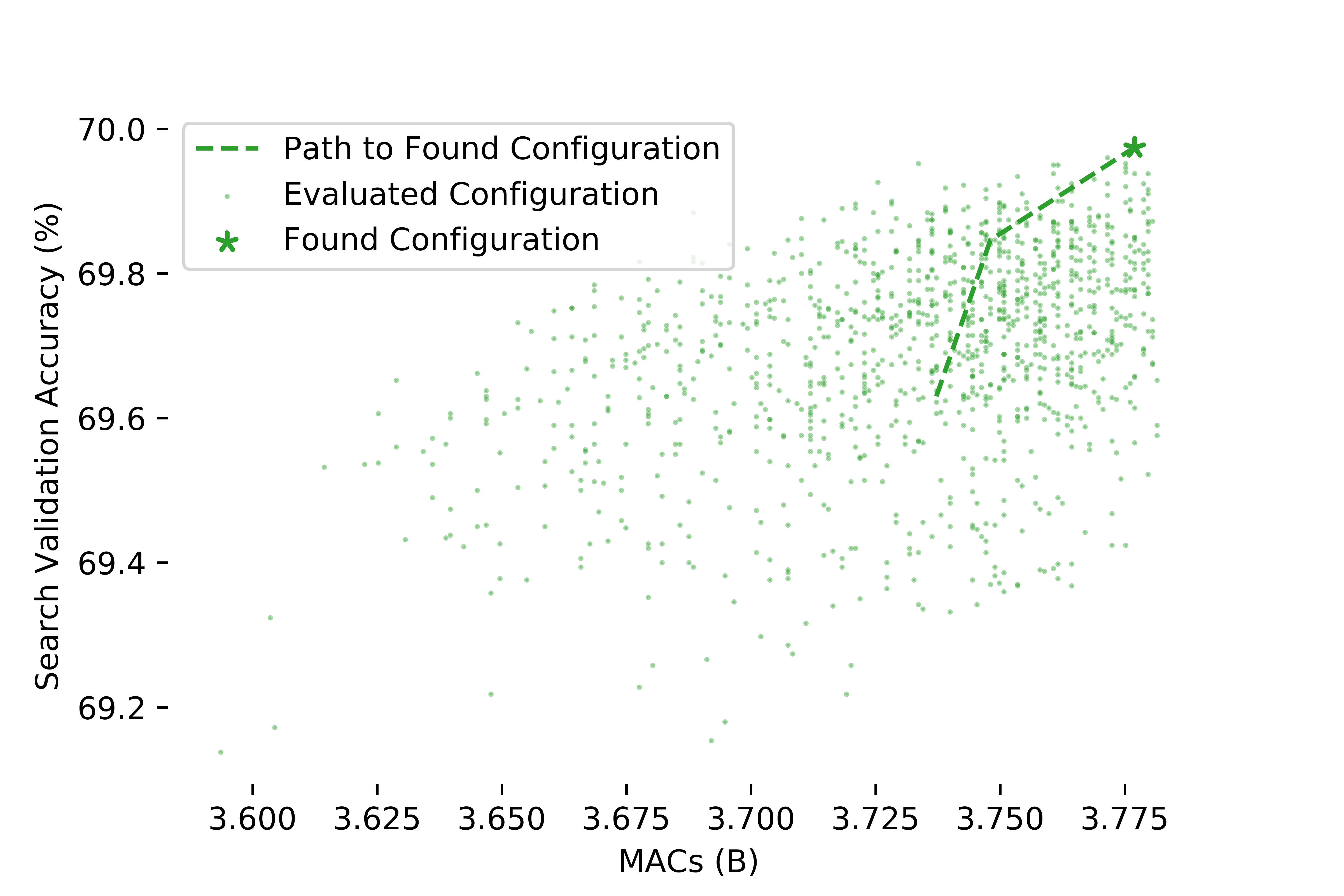}}
\subfloat{\includegraphics[width=0.45\textwidth]{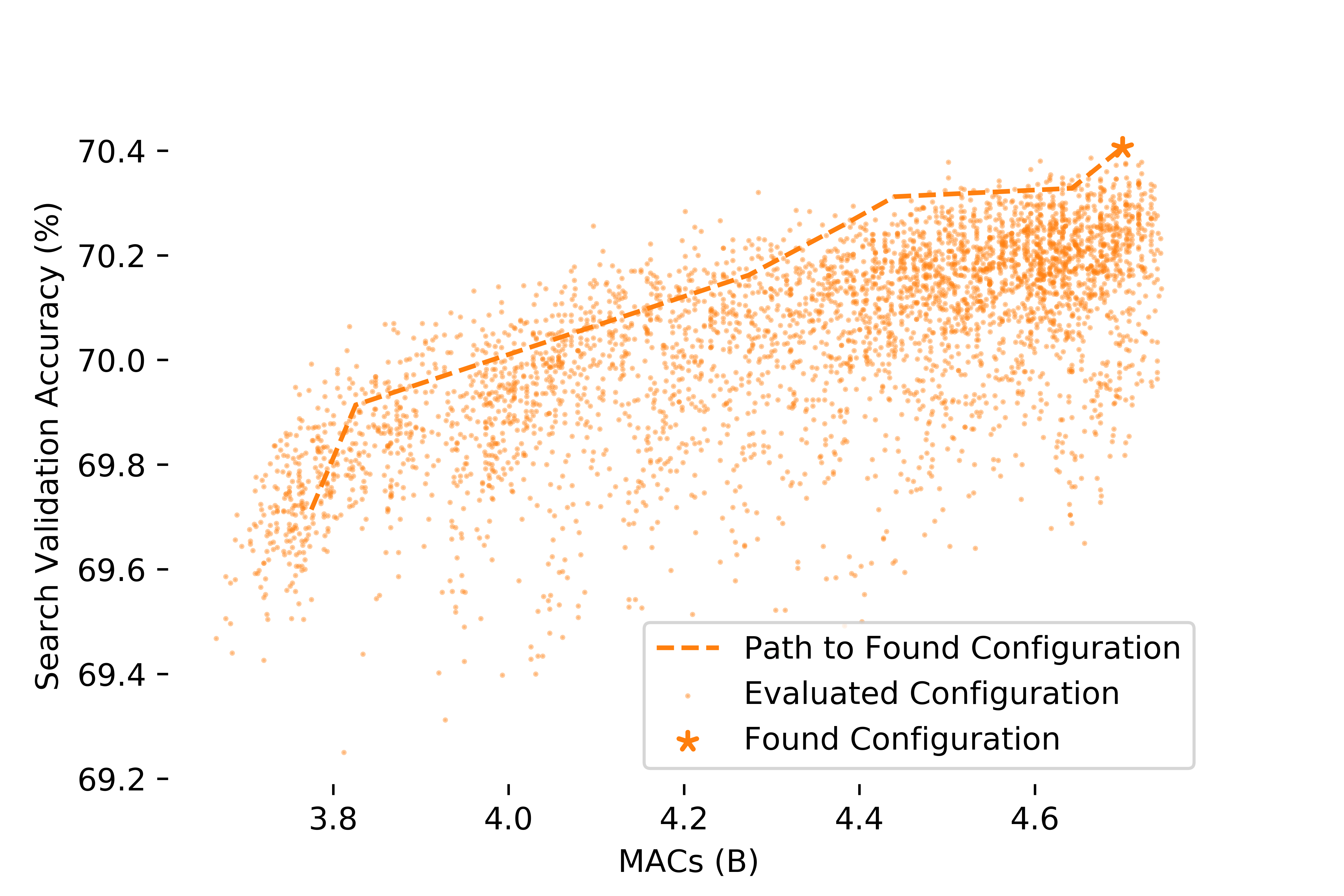}}
\end{center}
\caption{Visualisation of the VGG-16 breadth-first search on ImageNet. (\textbf{Left}) and (\textbf{Right}) are the searches against the \third{4s} and \fifth{16s} baselines, respectively.}
\label{fig:imgnet_search}
\end{figure}

We now move to a larger, more complex dataset in ImageNet. We perform the same GroSS decomposition and breadth-first search on conventional VGG-16 and ResNet-18 structures. We search against baseline configurations of uniform 4s and 16s. The results are listed in Table~\ref{tab:vgg_im_search} and we provide visualisation of the search on VGG-16 in Figure~\ref{fig:imgnet_search}.

\begin{table}[t]
    \centering
    \caption{Breadth-first search for VGG-16 and ResNet-18 on ImageNet. * denotes the configuration is using the decomposition structure from~\cite{wang2018deepsearch}}
    \resizebox{\columnwidth}{!}{%
    \begin{tabular}{ccccccc}
    \toprule
        \textbf{ImageNet}& & \multicolumn{2}{c}{\textbf{Accuracy}} & & \multicolumn{2}{c}{\textbf{$\Delta$MACs}} \\
        \textbf{Configuration} & \textbf{MACs} & GroSS & True & $\Delta$Acc. & Total & G.Conv \\ \midrule
        \textbf{VGG-16} (Full) & 15.49B & - & 71.59 & - & - & - \\ \addlinespace[0.25em]
        % \textit{Baseline: 32s} & \textit{2.82B} & \textit{} & \textit{70.96} & - & -\\
        % Found: & \textbf{} & \textbf{} & \textbf{} & $\downarrow$ -\% & $\uparrow$ - \\ \hline
    
        \textit{\fifth{16}\fifth{16}\fifth{16}\fifth{16}\fifth{16}\fifth{16}\fifth{16}\fifth{16}\fifth{16}\fifth{16}\fifth{16}\fifth{16}} & \textit{4.75B} & \textit{70.25} & \textit{70.77} & - & - & -\\
        \first{1}\sixth{32}\sixth{32}\sixth{32}\fifth{16}\first{1}\sixth{32}\fifth{16}\sixth{32}\third{4}\sixth{32}\fifth{16} & \textbf{4.70B} & \textbf{70.40} & \textbf{70.82} & $\mathbf{\uparrow 0.04}$ & $\mathbf{\downarrow0.99\%}$  &  $\mathbf{\downarrow3.65\%}$ \\ \addlinespace[0.25em]
    
        \textit{\third{4}\third{4}\third{4}\third{4}\third{4}\third{4}\third{4}\third{4}\third{4}\third{4}\third{4}\third{4}} & \textit{3.78B} & \textit{69.73} & \textit{70.51} & - & - & - \\
        \first{1}\first{1}\third{4}\third{4}\third{4}\third{4}\sixth{32}\fifth{16}\first{1}\first{1}\sixth{32}\first{1} & \textbf{3.78B}& \textbf{69.97} & \textbf{70.63} & $\mathbf{\uparrow 0.12}$ & $\mathbf{\downarrow0.14\%}$  & $\mathbf{\downarrow1.69\%}$ \\ \addlinespace[0.25em]
        
        \textit{\first{1}\first{1}\first{1}\first{1}\first{1}\first{1}\first{1}\first{1}\first{1}\first{1}\first{1}\first{1}} & \textit{3.54B} & \textit{68.98} & \textit{70.28} & - & - & -\\
        
    \cmidrule{1-1}
        
        \cite{wang2018deepsearch}: \textit{\fourth{11}\fourth{10}\fourth{14}\fourth{9}\fifth{15}\fifth{16}\fifth{16}\sixth{29}\sixth{33}\seventh{56}\seventh{56}\seventh{56}}* & \textit{1.16B} & \textit{62.64} & \textit{66.85} & - & - & - \\
        
        \fourth{11}\fourth{10}\fourth{14}\fourth{9}\fifth{15}\sixth{32}\seventh{64}\seventh{58}\second{3}\seventh{56}\third{7}\third{7}* & \textbf{1.16B} & \textbf{63.11} & \textbf{67.22} & $\mathbf{\uparrow 0.37}$ & $\mathbf{\downarrow0.39\%}$ & $\mathbf{\downarrow4.89\%}$ \\

    \midrule
        \textbf{ResNet-18} (Full) & 1.82B & - & 69.76 & - & - & -\\
        \addlinespace[0.25em]
        \textit{Baseline \fifth{16s}} & \textit{738M} & \textit{60.77} & \textit{65.80} & - & - & -\\
        \fifth{16}\fifth{16}\third{4}\sixth{32}\sixth{32}\sixth{32}\sixth{32}\sixth{32}\fifth{16}\sixth{32}\sixth{32}\sixth{32}\fifth{16}\third{4}\sixth{32}\fifth{16} & \textbf{715M} & \textbf{61.25} & \textbf{65.84} & $\mathbf{\uparrow 0.04}$ & $\mathbf{\downarrow3.18\%}$ & $\mathbf{\downarrow11.5\%}$ \\
        \addlinespace[0.25em]
        
        \textit{Baseline \third{4s}} & \textit{586M} & \textit{60.02} & \textit{65.46} & - & - & -\\
        \first{1}\third{4}\third{4}\first{1}\fifth{16}\third{4}\first{1}\third{4}\sixth{32}\third{4}\third{4}\third{4}\fifth{16}\third{4}\third{4}\third{4}& \textbf{585M} & \textbf{60.31} & \textbf{65.44} & $\mathbf{\uparrow 0.18}$ & $\mathbf{\downarrow0.08\%}$ & $\mathbf{\downarrow0.88\%}$\\
        \addlinespace[0.25em]
        
        \textit{\first{1}\first{1}\first{1}\first{1}\first{1}\first{1}\first{1}\first{1}\first{1}\first{1}\first{1}\first{1}} & \textit{547M} & \textit{58.61} & \textit{65.16} & - & - & -\\

    \bottomrule
    \end{tabular}
    }
    \label{tab:vgg_im_search}
\end{table}

We also include results for an alternative decomposition structure and group configuration identical to that used in~\cite{wang2018deepsearch}, which we detail in Appendix~\ref{app:wang}. This decomposition structure aggressively reduces widths in bottleneck layers to achieve a large compression ratio. In our search, we are able to show that the original configuration of groups within this structure is not optimal, with our found configuration leading to a significant improvement in accuracy as well as a slight speed up.

The results show that exploration of group selection with GroSS generalises well across datasets and architectures. In every search performed, we found configurations that met the objective of increased accuracy with lower inference cost.

\begin{figure}
\begin{center}
\includegraphics[width=0.70\textwidth]{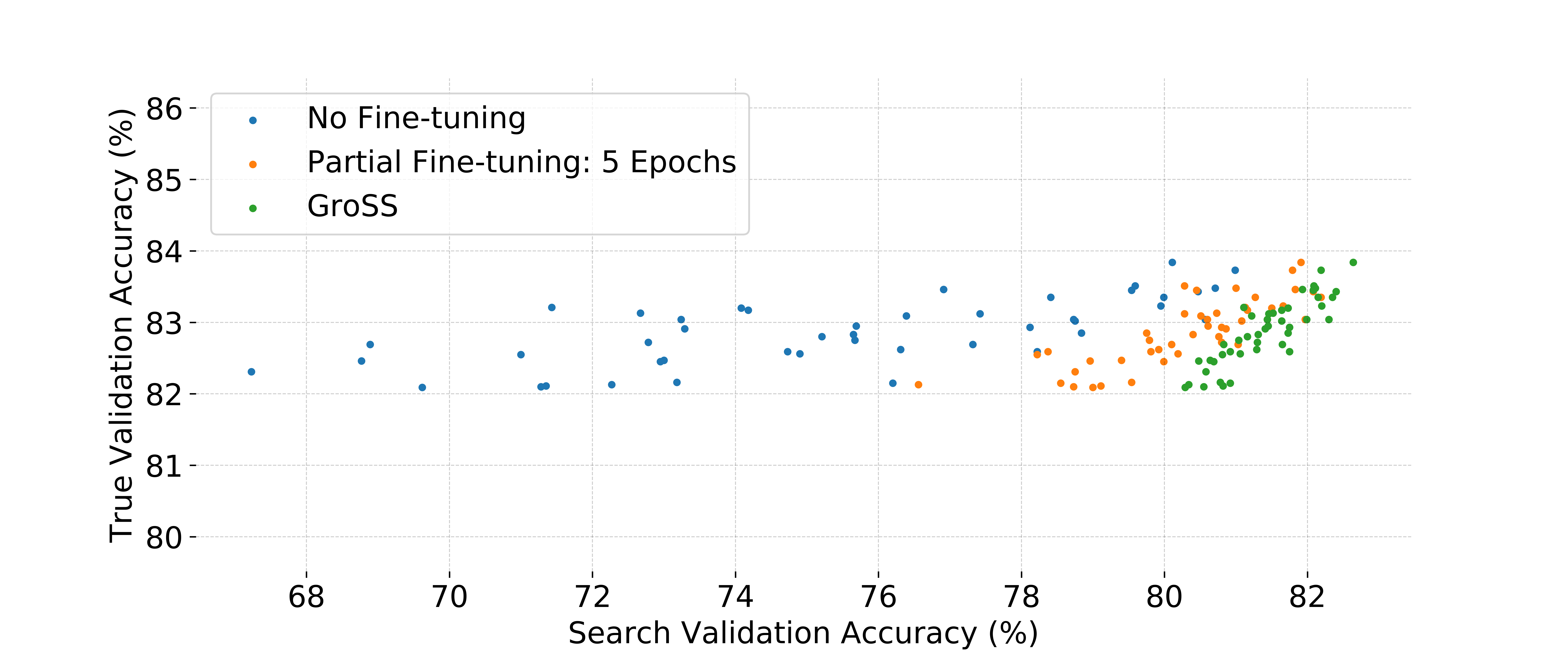}
\end{center}
\caption{Search-space vs true validation accuracy for our 4-layer network on CIFAR-10. Here we plot the accuracy of 45 random configurations of our 4-layer network for 3 different methods of obtaining a search-space. The accuracy of each configuration is plotted against its true validation accuracy.}
\label{fig:truevsearch}
\end{figure}

\subsection{GroSS Vs. Conventional Fine-tuning}
In this section, we justify the need for GroSS by evaluating it against using a partial fine-tuning strategy for each individual configuration. For this, we select 45 random group-size configurations of our 4-layer network and fine-tine according to our individual schedule, which is outlined in Sec.~\ref{sec:indconf}, giving us a true validation accuracy for each configuration. We can then evaluate the validation accuracy of these same configurations in our GroSS search-space. This procedure allows us to visualise how representative GroSS is of the true accuracy.

For comparison, we also include the validation accuracy of the same configurations with no fine-tuning, as well as with a shortened schedule of 5 epochs. The partial fine-tune could be considered a reasonable solution to reducing the burden of training while performing a configuration search.

We visualise the search-space against the true validation accuracies in Figure~\ref{fig:truevsearch}. Qualitatively, it can be seen that the validation accuracies produced by GroSS produce a significantly more consistent search space. The points appear to be more tightly distributed and closer to the ideal distribution ($y = x$). To measure this quantitatively, we compute the top-5 average precision of the search spaces. We simulate searches across the entire range of configurations by evaluating the average precision at multiple slices through the search-space. This allows for comparison across the space, not just the most accurate group configurations.

\begin{table}[t]
    \centering
    \caption{Average precision across the range of the search space. We compute the average precision using the top 5 true validation accuracies as positive recalls. ``$X\downarrow$" refers to the average precision computed after the top-X configurations have been removed from the search}
    \begin{tabular}{lccccc}
    \toprule
        \textbf{Fine-tune} & \multicolumn{5}{c}{\textbf{Average Precision (Top 5)}} \\
        \textbf{Strategy} & \textbf{All} & $\mathbf{10\downarrow}$ & $\mathbf{20\downarrow}$ & $\mathbf{30\downarrow}$ & \textbf{Mean} \\ \midrule
        No Fine-tuning & 43.0 & 43.8 & 86.3 & 69.8 & 60.7\\
        Partial Fine-tuning &  35.2 & \textbf{64.0} & 42.5 & 66.4 & 52.0 \\
        GroSS & \textbf{44.1} & 63.8 & \textbf{94.3} & \textbf{82.5} & \textbf{71.2}\\
    \bottomrule
    \end{tabular}

    \label{tab:avpre}
\end{table}

Table~\ref{tab:avpre} lists the results of this average precision computation. GroSS is consistently as good or better than no fine-tuning and the partial schedule at each slice. This leads to a significant improvement in search performance across the range of configurations which is highlighted by the mean average precision.

When making the comparison between GroSS and a partial training strategy, it is worth considering the computational requirements of each. Running inference for a new configuration in either of the conventional decompositions requires a new network to be initialised, and weights to be loaded. However, since group-sizes are handled dynamically within a GroSS decomposition, switching between them is essentially free, with no structure change or weight loading. This leads to GroSS having a significant speed improvement for running inference (7s vs 287s) over the 45 configurations. This only increases with more configurations tested. For example, the inference for the exhaustive search on the GroSS decomposition of the 4-layer network took only 9s for 252 configurations. Similarly, the total number of training epochs for the partial training strategy increase linearly with the number of configurations, but remain constant for GroSS. With larger search-spaces, such as those visualised in Figure~\ref{fig:imgnet_search}, the accuracy and performance benefits of GroSS combine to make grouped architecture search feasible where it might not have been before.

\section{Conclusions}

In this paper, we have presented GroSS, a series BTD factorisation which allows for the dynamic assignment and simultaneous training of differing numbers of groups within a layer. We have shown how GroSS-decomposed layers can be combined to train an entire grouped convolution search space at once. We confirmed the value of these configurations through an exhaustive search and a breadth-first search. We further demonstrate that, without GroSS, these searches would be less effective and dramatically less efficient.

\subsubsection*{Acknowledgements}

We gratefully acknowledge the European Commission Project Multiple-actOrs Virtual Empathic CARegiver for the Elder (MoveCare) for financially supporting the authors for this work.

% \clearpage
\bibliographystyle{splncs04}
\bibliography{egbib}

\begin{thebibliography}{10}
\providecommand{\url}[1]{\texttt{#1}}
\providecommand{\urlprefix}{URL }
\providecommand{\doi}[1]{https://doi.org/#1}

\bibitem{brock2017smash}
Brock, A., Lim, T., Ritchie, J.M., Weston, N.: {SMASH}: One-shot model
  architecture search through hypernetworks. arXiv Preprint arXiv:1708.05344
  (2017)

\bibitem{chen2018sharing}
Chen, Y., Jin, X., Kang, B., Feng, J., Yan, S.: Sharing residual units through
  collective tensor factorization to improve deep neural networks. In: IJCAI.
  pp. 635--641 (2018)

\bibitem{chollet2017xception}
Chollet, F.: Xception: Deep learning with depthwise separable convolutions. In:
  Proceedings of the IEEE Conference on Computer Vision and Pattern
  Recognition. pp. 1251--1258 (2017)

\bibitem{de2008decompositions}
De~Lathauwer, L.: Decompositions of a higher-order tensor in block terms—part
  {II}: Definitions and uniqueness. SIAM Journal on Matrix Analysis and
  Applications  \textbf{30}(3),  1033--1066 (2008)

\bibitem{denton2014exploiting}
Denton, E.L., Zaremba, W., Bruna, J., LeCun, Y., Fergus, R.: Exploiting linear
  structure within convolutional networks for efficient evaluation. In:
  Advances in Neural Information Processing Systems. pp. 1269--1277 (2014)

\bibitem{elthakeb2018releq}
Elthakeb, A.T., Pilligundla, P., Yazdanbakhsh, A., Kinzer, S., Esmaeilzadeh,
  H.: {R}e{L}e{Q}: A reinforcement learning approach for deep quantization of
  neural networks. arXiv Preprint arXiv:1811.01704  (2018)

\bibitem{he2015delving}
He, K., Zhang, X., Ren, S., Sun, J.: Delving deep into rectifiers: Surpassing
  human-level performance on {I}mage{N}et classification. In: Proceedings of
  the IEEE International Conference on Computer Vision. pp. 1026--1034 (2015)

\bibitem{he2016deep}
He, K., Zhang, X., Ren, S., Sun, J.: Deep residual learning for image
  recognition. In: Proceedings of the IEEE Conference on Computer Vision and
  Pattern Recognition. pp. 770--778 (2016)

\bibitem{howard2017mobilenets}
Howard, A.G., Zhu, M., Chen, B., Kalenichenko, D., Wang, W., Weyand, T.,
  Andreetto, M., Adam, H.: {M}obile{N}ets: Efficient convolutional neural
  networks for mobile vision applications. arXiv Preprint arXiv:1704.04861
  (2017)

\bibitem{ioannou2017deep}
Ioannou, Y., Robertson, D., Cipolla, R., Criminisi, A.: Deep roots: Improving
  cnn efficiency with hierarchical filter groups. In: Proceedings of the IEEE
  Conference on Computer Vision and Pattern Recognition. pp. 1231--1240 (2017)

\bibitem{jaderberg2014speeding}
Jaderberg, M., Vedaldi, A., Zisserman, A.: Speeding up convolutional neural
  networks with low rank expansions. arXiv Preprint arXiv:1405.3866  (2014)

\bibitem{kim2015compression}
Kim, Y.D., Park, E., Yoo, S., Choi, T., Yang, L., Shin, D.: Compression of deep
  convolutional neural networks for fast and low power mobile applications.
  arXiv Preprint arXiv:1511.06530  (2015)

\bibitem{kossaifi2019tensorly}
Kossaifi, J., Panagakis, Y., Anandkumar, A., Pantic, M.: Tensorly: Tensor
  learning in python. The Journal of Machine Learning Research  \textbf{20}(1),
   925--930 (2019)

\bibitem{krizhevsky2014cifar}
Krizhevsky, A., Nair, V., Hinton, G.: The {CIFAR}-10 dataset. Online:
  Http://www. Cs. Toronto. Edu/Kriz/Cifar. HTML  \textbf{55} (2014)

\bibitem{krizhevsky2012imagenet}
Krizhevsky, A., Sutskever, I., Hinton, G.E.: {I}mage{N}et classification with
  deep convolutional neural networks. In: Advances in Neural Information
  Processing Systems. pp. 1097--1105 (2012)

\bibitem{lebedev2014speeding}
Lebedev, V., Ganin, Y., Rakhuba, M., Oseledets, I., Lempitsky, V.: Speeding-up
  convolutional neural networks using fine-tuned {CP}-decomposition. arXiv
  Preprint arXiv:1412.6553  (2014)

\bibitem{li2019random}
Li, L., Talwalkar, A.: Random search and reproducibility for neural
  architecture search. arXiv Preprint arXiv:1902.07638  (2019)

\bibitem{liu2018darts}
Liu, H., Simonyan, K., Yang, Y.: {DARTS}: Differentiable architecture search.
  arXiv Preprint arXiv:1806.09055  (2018)

\bibitem{nakajima2013global}
Nakajima, S., Sugiyama, M., Babacan, S.D., Tomioka, R.: Global analytic
  solution of fully-observed variational {B}ayesian matrix factorization.
  Journal of Machine Learning Research  \textbf{14}(Jan),  1--37 (2013)

\bibitem{paszke2019pytorch}
Paszke, A., Gross, S., Massa, F., Lerer, A., Bradbury, J., Chanan, G., Killeen,
  T., Lin, Z., Gimelshein, N., Antiga, L., Desmaison, A., Kopf, A., Yang, E.,
  DeVito, Z., Raison, M., Tejani, A., Chilamkurthy, S., Steiner, B., Fang, L.,
  Bai, J., Chintala, S.: Pytorch: An imperative style, high-performance deep
  learning library. In: Wallach, H., Larochelle, H., Beygelzimer, A.,
  d'Alch\'{e} Buc, F., Fox, E., Garnett, R. (eds.) Advances in Neural
  Information Processing Systems 32, pp. 8024--8035. Curran Associates, Inc.
  (2019),
  \url{http://papers.neurips.cc/paper/9015-pytorch-an-imperative-style-high-performance-deep-learning-library.pdf}

\bibitem{sandler2018mobilenetv2}
Sandler, M., Howard, A., Zhu, M., Zhmoginov, A., Chen, L.C.: {M}obile{N}et{V}2:
  Inverted residuals and linear bottlenecks. In: Proceedings of the IEEE
  Conference on Computer Vision and Pattern Recognition. pp. 4510--4520 (2018)

\bibitem{sifre2014rigid}
Sifre, L., Mallat, S.: Rigid-motion scattering for image classification  (2014)

\bibitem{simonyan2014very}
Simonyan, K., Zisserman, A.: Very deep convolutional networks for large-scale
  image recognition. arXiv Preprint arXiv:1409.1556  (2014)

\bibitem{tan2019mnasnet}
Tan, M., Chen, B., Pang, R., Vasudevan, V., Sandler, M., Howard, A., Le, Q.V.:
  {M}nas{N}et: Platform-aware neural architecture search for mobile. In:
  Proceedings of the IEEE Conference on Computer Vision and Pattern
  Recognition. pp. 2820--2828 (2019)

\bibitem{tan2019efficientnet}
Tan, M., Le, Q.V.: {E}fficient{N}et: Rethinking model scaling for convolutional
  neural networks. arXiv Preprint arXiv:1905.11946  (2019)

\bibitem{tucker1966some}
Tucker, L.R.: Some mathematical notes on three-mode factor analysis.
  Psychometrika  \textbf{31}(3),  279--311 (1966)

\bibitem{vanhoucke2011improving}
Vanhoucke, V., Senior, A., Mao, M.Z.: Improving the speed of neural networks on
  {CPU}s  (2011)

\bibitem{wang2018deepsearch}
Wang, P., Hu, Q., Fang, Z., Zhao, C., Cheng, J.: {D}eep{S}earch: A fast image
  search framework for mobile devices. ACM Transactions on Multimedia
  Computing, Communications, and Applications (TOMM)  \textbf{14}(1), ~6 (2018)

\bibitem{wu2019fbnet}
Wu, B., Dai, X., Zhang, P., Wang, Y., Sun, F., Wu, Y., Tian, Y., Vajda, P.,
  Jia, Y., Keutzer, K.: {FBN}et: Hardware-aware efficient convnet design via
  differentiable neural architecture search. In: Proceedings of the IEEE
  Conference on Computer Vision and Pattern Recognition. pp. 10734--10742
  (2019)

\bibitem{xie2017aggregated}
Xie, S., Girshick, R., Doll{\'a}r, P., Tu, Z., He, K.: Aggregated residual
  transformations for deep neural networks. In: Proceedings of the IEEE
  Conference on Computer Vision and Pattern Recognition. pp. 1492--1500 (2017)

\bibitem{zhang2018shufflenet}
Zhang, X., Zhou, X., Lin, M., Sun, J.: {S}huffle{N}et: An extremely efficient
  convolutional neural network for mobile devices. In: Proceedings of the IEEE
  Conference on Computer Vision and Pattern Recognition. pp. 6848--6856 (2018)

\bibitem{zoph2018learning}
Zoph, B., Vasudevan, V., Shlens, J., Le, Q.V.: Learning transferable
  architectures for scalable image recognition. In: Proceedings of the IEEE
  Conference on Computer Vision and Pattern Recognition. pp. 8697--8710 (2018)

\end{thebibliography}

\clearpage
\appendix
\section{Appendix}
\label{appendix}

\subsection{Network Definitions}
\label{app:network}
In this section, we will provide the explicit definitions of the networks used for our experiments. In Table~\ref{tab:4layerstruct}, we detail the structure of our 4-layer network. Table~\ref{tab:vggclassstruct} lists the non-standard classifier structure used for VGG-16 on CIFAR-10. All convolutional and fully-connected layers are followed by a ReLU non-linearity, with the exception of the final fully-connected layers in the classifiers.

\begin{table}
    \centering
    \caption{Architecture of our 4-layer network. Each convolution has a $3\times3$ kernel and is followed by a ReLU non-linearity and a $2\times2$ max pooling layer}
    \begin{tabular}{ccccc} \toprule
        \textbf{Conv 1} & \textbf{Conv 2} & \textbf{Conv 3} & \textbf{Conv 4} & \textbf{Classifier} \\ \midrule
        conv($3\xrightarrow{}32$) & conv($32\xrightarrow{}32$) & conv($32\xrightarrow{}64$) & conv($64\xrightarrow{}64$) & fc(256$\xrightarrow{}$256) \\
         &  &  &  & fc(256$\xrightarrow{}$10) \\
        \bottomrule
    \end{tabular}
    \label{tab:4layerstruct}
\end{table}

\begin{table}
    \centering
    \caption{VGG-16 classifier structure for CIFAR-10 and ImageNet}
    \begin{tabular}{cc} \toprule
        \textbf{Dataset} & \textbf{Layers} \\ \midrule
        \multirow{2}{*}{CIFAR-10\space\space} & fc(512$\xrightarrow{}$512) \\ & fc(512$\xrightarrow{}$10)  \\ \addlinespace[0.75em]
         & fc(25088$\xrightarrow{}$4096) \\ImageNet\space\space & fc(4096$\xrightarrow{}$4096) \\ & fc(4096$\xrightarrow{}$1000) \\ \bottomrule
    \end{tabular}
    \label{tab:vggclassstruct}
\end{table}

\subsection{Training From Scratch}
\label{app:training}
\textbf{4-layer Network.} Convolutional weights in the network are initialised with the He initialisation~\cite{he2015delving} in the ``fan out" mode with a ReLU non-linearity. The weights of the fully-connected layers are initialised with a zero-mean, 0.01-variance normal distribution. All bias terms in the network are initialised to 0. The network is trained from scratch on our CIFAR-10 training split for 100 epochs using stochastic gradient descent (SGD). We adopt a initial learning rate of 0.1 and momentum of 0.9. The learning rate is decayed by a factor of 0.1 after 50 and 75 epochs. We train the network 5 times and use the weights with median accuracy for further experiments.

\textbf{VGG-16.} For our CIFAR-10 variant of VGG-16, the weights are initialised with identical strategy to the 4-layer network. We train this full network on CIFAR-10 for a total of 200 epochs, again using stochastic gradient descent. The initial learning rate is set to 0.05 and momentum to 0.9. The learning rate is decayed by a factor of 0.1 after 100 and 150 epochs. For ImageNet, we take the pretrained model from the Pytorch (Torchvision)~\cite{paszke2019pytorch} model zoo. Specifically, we take the variant without batch-normalisation layers.

\textbf{ResNet-18.} We again make use of the Torchvision model zoo, and use their ResNet-18 model trained on ImageNet.

\subsection{High-Compression Decomposition Structure}
\label{app:wang}
We recreate the exact structure used for VGG-16 acceleration in~\cite{wang2018deepsearch} with GroSS, which is listed in~\ref{tab:wang}. The constraint that is used in our other experiments, where bottlenecks should be constant width, is relaxed. Since the group-size must be a factor of both bottleneck dimensions (in and out), the bottleneck dimensions chosen by Wang~\etal do limit the choice of ranks in GroSS. We perform decomposition as with our other VGG-16 experiments, without any bells or whistles. We found that a longer fine-tuning schedule was required to best recover accuracy. Therefore, finetuning consists of 14 epochs with a learning rate of $5 \times 10^{-4}$ and decay after 8 and 12 epochs. This led to an accuracy consistent with~\cite{wang2018deepsearch}.
\begin{table}[h]
    \centering
    \caption{Decomposition structure as used in~\cite{wang2018deepsearch}. Group-sizes marked with * represent the original choice}
    \begin{tabular}{ccc} \toprule
         & Bottleneck & \\
        Layer & (in $\rightarrow$ out) & Group-sizes  \\ \midrule
        \verb^conv1_2^ & 11 $\rightarrow$ 18 & \fourth{11*}\\
        \verb^conv2_1^ & 10 $\rightarrow$ 24 & \second{5}\fourth{10*}\\
        \verb^conv2_2^ & 28 $\rightarrow$ 28 & \first{1}\third{7}\fourth{14*}\sixth{28}\\
        \verb^conv3_1^ & 36 $\rightarrow$ 48 & \second{3}\third{9*}\fifth{18}\\
        \verb^conv3_2^ & 60 $\rightarrow$ 48 & \fifth{15*}\sixth{30}\seventh{60}\\
        \verb^conv3_3^ & 64 $\rightarrow$ 56 & \fifth{16*}\sixth{32}\seventh{64}\\
        \verb^conv4_1^ & 64 $\rightarrow$ 100 & \fifth{16*}\fifth{32}\seventh{64}\\
        \verb^conv4_2^ & 116 $\rightarrow$ 100 & \sixth{29*}\seventh{58}\eigth{116}\\
        \verb^conv4_3^ & 132 $\rightarrow$ 132 & \second{3}\sixth{33*}\seventh{66}\\
        \verb^conv5_1^ & 224 $\rightarrow$ 224 & \third{7}\sixth{28}\seventh{56*}\eigth{112}\\
        \verb^conv5_2^ & 224 $\rightarrow$ 224 & \third{7}\sixth{28}\seventh{56*}\eigth{112}\\
        \verb^conv5_3^ & 224 $\rightarrow$ 224 & \third{7}\sixth{28}\seventh{56*}\eigth{112}\\
        \bottomrule
    \end{tabular}
    \label{tab:wang}
\end{table}

\end{document}